\documentclass[runningheads]{llncs}

\usepackage{eccv}
\usepackage{eccvabbrv}
\usepackage{graphicx}
\usepackage{booktabs}
\usepackage[accsupp]{axessibility} 
\usepackage{hyperref}
\usepackage{orcidlink}

\usepackage{amsmath}
\usepackage{enumitem}
\usepackage{booktabs}
\usepackage{xcolor}
\definecolor{mygrey}{cmyk}{0, 0, 0, 59}
\usepackage{multirow}
\usepackage{amssymb}
\usepackage{bm}
\usepackage{colortbl}
\usepackage{tabularx}
\usepackage{array}

\usepackage{algorithm} 
\usepackage{algpseudocode}
\usepackage{lipsum}
\usepackage{float}

\definecolor{lightgray}{rgb}{0.45, 0.45, 0.45}
\newcommand{\commentcolor}[1]{{\color{lightgray}{#1}}}



\begin{document}

\title{Soft Prompt Generation for Domain Generalization} 

\titlerunning{SPG for DG}

\author{
Shuanghao Bai\inst{1\star} \quad
Yuedi Zhang\inst{1\star} \quad
Wanqi Zhou\inst{1} \quad \\
Zhirong Luan\inst{2} \quad
Badong Chen\inst{1\dag}
}
\institute{
$^{1}$Institute of Artificial Intelligence and Robotics, Xi’an Jiaotong University, China \quad 
$^{2}$School of Electrical Engineering, Xi’an University of Technology, Xi'an, China \\
\email{\{baishuanghao, zyd993, zwq785915792\}@stu.xjtu.edu.cn, luanzhirong@xaut.edu.cn, chenbd@mail.xjtu.edu.cn}
}

\authorrunning{Bai et al.}

\maketitle

\renewcommand{\thefootnote}{}
\footnotetext[1]{$^\star$ Equal contribution.}
\footnotetext[2]{$^\dag$ Corresponding author.}

\begin{abstract}

Large pre-trained vision language models (VLMs) have shown impressive zero-shot ability on downstream tasks with manually designed prompt. 
To further adapt VLMs to downstream tasks, soft prompt is proposed to replace manually designed prompt, which undergoes fine-tuning based on specific domain data.
Prior prompt learning methods primarily learn a fixed prompt or residuled prompt from training samples. However, the learned prompts lack diversity and ignore information about unseen domains.
In this paper, we reframe the prompt learning framework from a generative perspective and propose a simple yet efficient method for the Domain Generalization (DG) task, namely \textbf{S}oft \textbf{P}rompt \textbf{G}eneration (SPG).
Specifically, SPG consists of a two-stage training phase and an inference phase. During the training phase, we introduce soft prompt label for each domain, aiming to incorporate the generative model domain knowledge. During the inference phase, the generator of the generative model is employed to obtain instance-specific soft prompts for the unseen target domain.
Extensive experiments on five domain generalization benchmarks of three DG tasks demonstrate that SPG achieves state-of-the-art performance.
The code is available at https://github.com/renytek13/Soft-Prompt-Generation-with-CGAN.

\keywords{Domain Generalization \and Visual Language Models \and Prompt Learning \and Generative Models}

\end{abstract}

\section{Introduction}
\label{sec:intro}

\begin{figure*}[ht]
  \centering
  \includegraphics[width=0.95\textwidth]{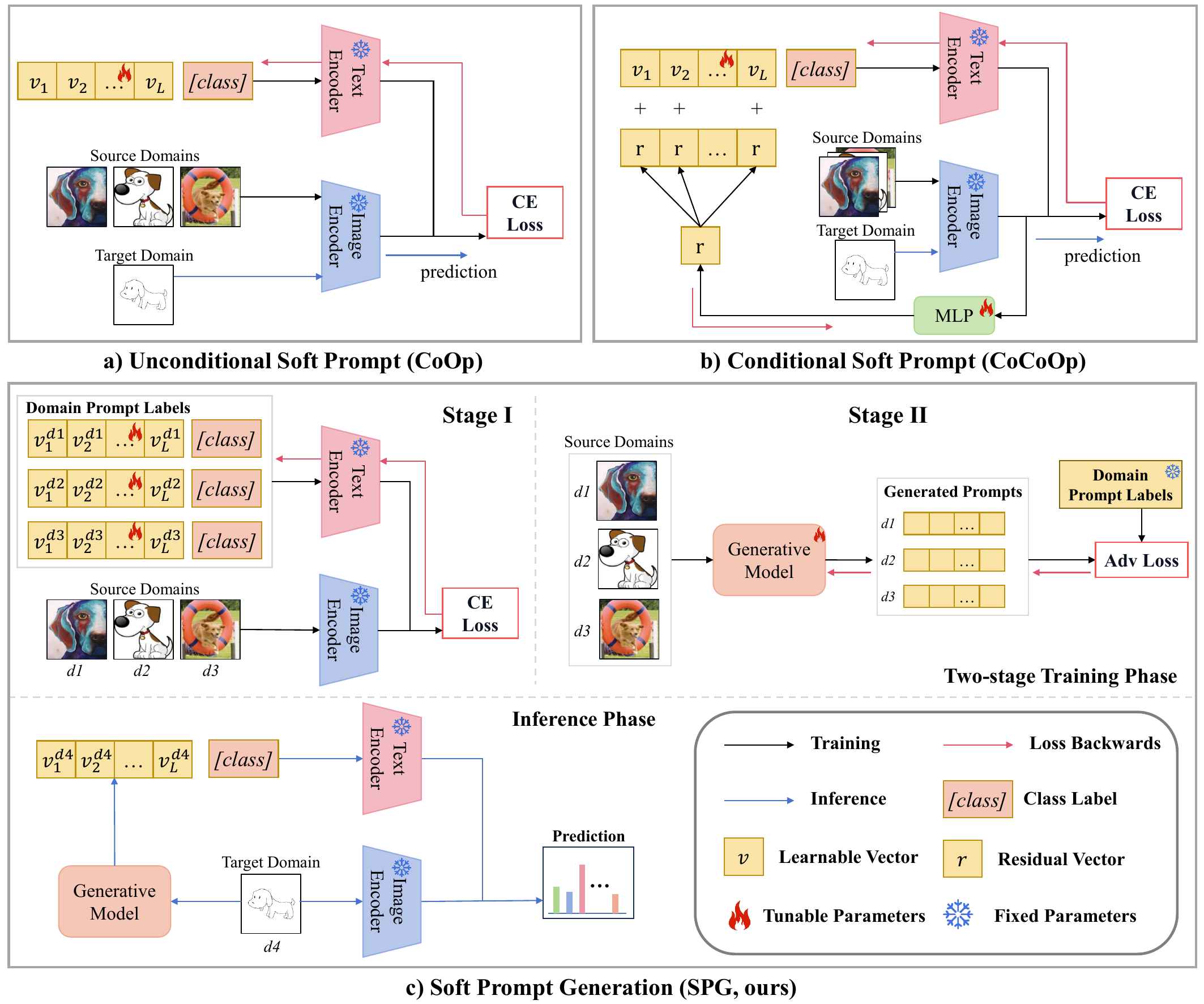}
  \caption{The difference between previous work and our work. We reframe the prompt learning framework from a generative perspective.  
  We exclusively rely on a generative model to directly produce soft prompts, ensuring their diversity. Essentially, we transfer the prompt's adaptability to the generative model by incorporating domain knowledge.
  }
  \label{fig:intro}
\end{figure*}

Large vision language models, such as CLIP~\cite{radford2021learning} and ALIGN~\cite{jia2021scaling}, have attracted significant attention owing to their effective adaptation to downstream tasks with manually designed prompts. However, manually designed prompts are not always optimal for domain-specific tasks. 
Instead of manually designed prompts, the soft prompt~\cite{zhou2022learning, qin2021learning, li2024dialogue} can be optimized in a data-driven manner through backpropagation.
The soft prompt, serving as a learning vector, is refined through fine-tuning with domain-specific data to better adapt to downstream tasks, including base-to-novel generalization~\cite{zhou2022learning, khattak2023maple} and domain adaptation~\cite{ge2023domain, bai2024prompt, wang2024clips}. 
Despite the progress made in prompt learning, generalization performance tends to decline significantly when facing distribution shifts.

Many efforts have been made to address the distribution shift problem by domain generalization (DG)~\cite{zhou2022domain, li2018deep}. 
The objective of DG is to train a model using data from one or multiple related but distinct source domains in a manner that enables the model to generalize effectively to any out-of-distribution (OOD) unseen target domain.
Due to the extensive learning in various domains, vision language models like CLIP exhibit a generalization performance that can even rival state-of-the-art traditional supervised learning algorithms. 
Taking a step, many works adopt prompt learning techniques~\cite{niu2022domain, zhou2022conditional, zhang2023domain, cho2023promptstyler, cheng2024disentangled} to enhance the generalization performance of CLIP. Niu et al.~\cite{niu2022domain} introduce domain bank to incorporate textual domain knowledge into soft prompts. Both Zhou et al.~\cite{zhou2022conditional} and Zhang et al.~\cite{zhang2023domain} establish a lightweight neural network (\textit{i.e.}, multilayer perceptron) to generate soft prompts conditioned on images with a residual or concatenation manner. 
However, these methods either lack diversity and visibility into the information from the target domain or rely on simple models to approximate the relationship between images and soft prompts. Consequently, the learned prompts may still fall short of being transferable.

As shown in Figure~\ref{fig:intro}, unlike the previous prompt learning methods, such as unconditional prompt learning~\cite{zhou2022learning, khattak2023maple} and conditional prompt learning~\cite{zhou2022conditional, derakhshani2023bayesian, zhang2023domain} methods, which typically rely on fixed prompts obtained from the training set for downstream tasks, or employ a straightforward multi-layer perceptron to learn a residual vector to enhance the richness of the fixed prompts, our approach takes a different direction. 
We reframe the prompt learning framework by adopting a generative perspective, marking the first integration of generative models into VLMs for prompt learning. 
We propose a new prompt learning paradigm \textbf{S}oft \textbf{P}rompt \textbf{G}eneration (SPG), which offers a straightforward yet effective solution for Domain Generalization. 
SPG is designed to exclusively harness a generative model for prompt generation, leveraging the model's inherent capability to encode domain and content knowledge directly into the generated prompts.

\textbf{Our Proposed SPG}: SPG method consists of a two-stage training phase and an inference phase.
Specifically, in the initial training phase, we introduce domain prompt labels, representing optimal prompts for each domain. Images and corresponding domain prompt labels are then input into a simple yet effective generative model, Conditional Generative Adversarial Net (CGAN)~\cite{mirza2014conditional}, to incorporate domain and content knowledge into the prompt generator model in the second training phase. As a result, domain knowledge is stored not in soft prompts but in the generative model.
During inference, the generator of the generative model is employed to obtain domain-specific soft prompts for the target domain data, enabling enhanced diversity and transferability for prompts.

Our main contributions are as follows:

\begin{itemize}
    \item To the best of our knowledge, we are the first to introduce the generative model into prompt learning in VLMs. Then, we propose a new paradigm of prompt tuning, namely \textbf{S}oft \textbf{P}rompt \textbf{G}eneration (SPG).
    \item We design a two-stage training phase to align the generative model with domain prompt labels. It incorporates domain knowledge into the generated prompts, enhancing the transferability across unseen domains.
    \item Extensive experiments on five datasets for three DG tasks demonstrate that the proposed SPG achieves state-of-the-art performance.
\end{itemize}

\section{Related Work}
\label{sec:rw}

\subsection{Domain Generalization}
Domain Generalization (DG) aims to train a model using data from one or multiple related yet different source domains, enabling the model to generalize effectively to any out-of-distribution (OOD) target domain.
Existing works mainly focus on learning domain-invariant features across one or multiple source domains.
One line of research focuses on domain alignment methods, primarily aiming to minimize moments~\cite{muandet2013domain}, KL Divergence~\cite{li2020domain}, and Maximum Mean Discrepancy~\cite{li2018domain} to learn domain-invariant features. 
Another line of work involves leveraging data augmentation to enrich images or their features, thereby contributing to the learning of invariant predictors~\cite{volpi2018generalizing, li2021simple}.
Additional strategies encompass adversarial learning~\cite{ganin2016domain}, ensemble learning~\cite{li2022simple}, meta-learning~\cite{li2018learning}, gradient operations~\cite{cha2021swad}, and more.
Recently, large vision language models (VLMs) such as CLIP~\cite{radford2021learning} have been applied to various downstream tasks, demonstrating their potential for remarkable generalization performance. 
One of the most efficient fine-tuning paradigms for VLMs is prompt learning. Building upon the success of prompt learning, our work delves deeper into exploring this highly efficient paradigm. 
We propose a novel prompt learning framework that departs from the previous prompt learning methods of utilizing fixed prompts~\cite{zhou2022learning} or residualed prompts~\cite{zhou2022conditional} directly during inference. Instead, our framework leverages a generative model to dynamically obtain soft prompts for the inference process, introducing a new paradigm of prompt tuning.

\subsection{Prompt Learning for Vision Language Models}
Prompt learning for VLMs, also referred to as prompt tuning, aims to tune the model on domain-specific downstream tasks by only training a small set of parameters which might be a set of newly added parameters with the input.
CoOp~\cite{zhou2022learning} firstly introduces soft prompt in VLMs and demonstrates a suitable prompt that can improve performance for the image recognition task. 
CoCoOp~\cite{zhou2022conditional} solves the overfitting problem of CoOp with conditioned prompt learning, i.e., residualed prompt, ensuring the diversity of prompt. 
DAPL~\cite{ge2023domain} and PDA~\cite{bai2024prompt} introduce domain labels and domain alignment into prompt learning for domain adaptation, respectively. 
For DG problem, CAE~\cite{niu2022domain} aims to obtain domain-unified text representation with domain bank. DPL~\cite{zhang2023domain} generates prompts through MLP to add domain-specific features to the prompt template.
Different from previous prompt learning methods in VLMs, we reframe the prompt learning framework from a generative perspective, i.e., exclusively relying on a generative model to dynamically produce soft prompts. 
For the DG setting, we further introduce a two-stage training paradigm and domain prompt labels to effectively embed domain knowledge within the generative model.

\section{Preliminaries}
\label{sec:pre}

\subsection{Problem Setup of Domain Generalization}

In DG setting, there are $M$ source domains $D_s={\{D^1_s, D^2_s, ..., D^M_s}\}$ and $N$ target domains $D_t={\{D^1_t, D^2_t, ..., D^N_t}\}$, all of which follow different distributions.
For each source domain, $D^i_s=\{{({x}^{i}_{j}, y^{i}_{j})}\}^{n_i}_{j=1}$ where $n_i$ denotes the size of samples in $D^i_s$ and $(x^{i}_{j}, y^{i}_{j})$ denotes the pair of input and target label for the $j$th sample in the $i$th domain. Then we denote the input space as $X$ and denote the label set as $Y$. 
Assuming that all domains share the same label space, previous methods mainly focus on learning a domain-invariant model to map $F: {X} \rightarrow Y$ from images to labels, with the aspiration that this mapping can generalize to unseen target domains.
With the advent of VLMs, prompt learning methods are proposed to incorporate soft prompts $V$ into the input, and the mapping is rephrased as $F: \{{X, V}\} \rightarrow Y$. 
In contrast to previous methods that involved fine-tuning the model, prompt learning methods center on the fine-tuning of the prompt $V$.
Unlike these methods above, we introduce a new prompt learning paradigm that solely relies on a generative model $G$ to produce soft prompts directly. Thus, the mapping is rephrased as $F: \{{X, G(X)}\} \rightarrow Y$.
Our goal is to learn a generative model that can capture both domain-invariant and domain-specific features to produce generalized prompts for unseen target domains dynamically.

\subsection{Prompt Learning Methods in Generalization}

Contrastive Language-Image Pre-Training (CLIP)~\cite{radford2021learning} model is pre-trained on 400 million image-text pairs collected from the internet with contrastive learning. 
It consists of an image encoder $f$ and a text encoder $g$, which encodes images and corresponding natural language descriptions, respectively.

\noindent \textbf{Zero-shot CLIP} directly incorporates a template text description as a prompt into the text encoder,  such as "a photo of a [CLASS]" where [CLASS] denotes the class token. 
The image features and text features $w$ of manually designed prompts are extracted from the image encoder and text encoder, respectively.
The prediction of the class of the image is 
$\hat{y}=\underset{k}{\arg \max }\left\langle f(\mathbf{x}), w_k\right\rangle$, where $\langle\cdot,\cdot\rangle$ denotes the cosine similarity.

\noindent \textbf{CoOp}~\cite{zhou2022learning} introduces a set of continuous learnable context vectors $v$ concatenated with the template prompt $c$, namely soft prompt, then the $i$th class of text prompt ${t}^i$ can be defined as ${t}^i=[v,c^i]$. Therefore the whole framework can be updated through the frozen CLIP model via tuning these prompts with cross-entropy loss $\mathcal{L}_{ce} = -\mathbb{E}_y\left[\sum_i y_i \log \left(\hat{y}_i\right | \mathbf{x}_i, t_i)\right]$, where $\hat{y}_i$ denotes $i$th element of the model's predicted probability distribution.
However, CoOp learns a fixed prompt from training samples, which may lead to overfitting to the training distribution and degraded performance on test distribution. 

\noindent \textbf{CoCoOp}~\cite{zhou2022conditional} \textbf{and DPL}~\cite{zhang2023domain} attempt to overcome distribution shifts by learning an instance-specific continuous prompt that is conditioned on the input image with an MLP layer $\phi$.
Both of them learn an image-conditional vector $r=\phi(f(\mathbf{x}))$, which is then either added for CoCoOp with the learnable context vectors $v$ or concatenated for DPL. Then the $i$th class of text prompt ${t}^i$ can be defined as ${t}^i=[v+r,c^i]$ or ${t}^i=[r,c^i]$. By tuning the prompts and MLP layer, the loss is formulated as $\mathcal{L}_{ce} = -\mathbb{E}_y\left[\sum_i y_i \log \left(\hat{y}_i\right | \mathbf{x}_i, t_i, \phi)\right]$.

\section{Method}
\label{sec:method}

\begin{figure*}[ht]
  \centering
  \includegraphics[width=0.9\textwidth]{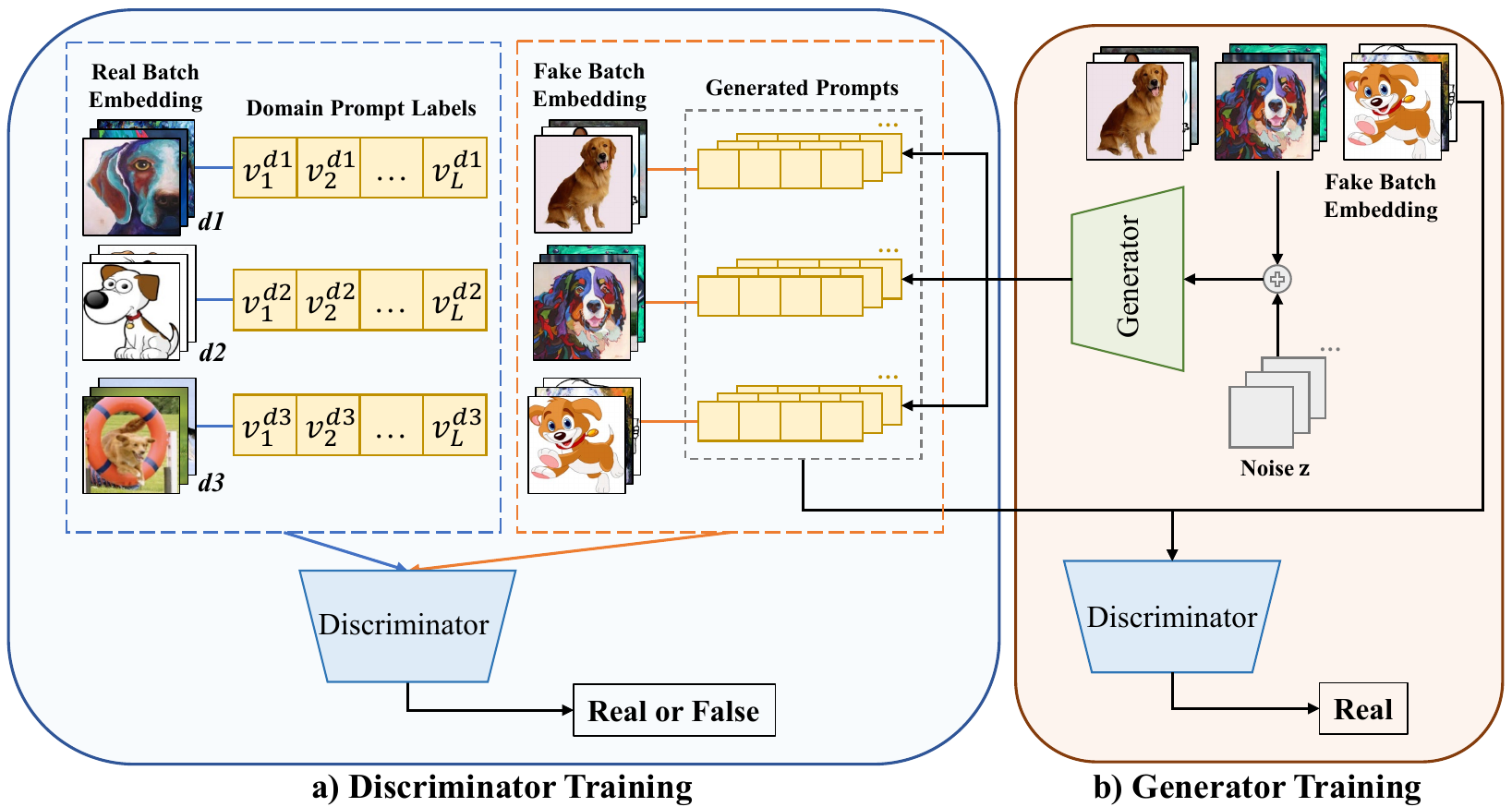}
  \caption{The design of the second stage of the training phase. The condition generative adversarial net is the backbone of the generative model. The generator is guided by images to produce prompts. Meanwhile, the discriminator evaluates the authenticity of the prompt labels and the generated prompts with image data.
  }
  \label{fig:stage2}
\end{figure*}

Different from previous prompt learning methods, we abandon the training paradigm of the fixed prompt and residualed prompt. Instead, we introduce a novel generative perspective to prompt learning, namely \textbf{S}oft \textbf{P}rompt \textbf{G}eneration (SPG).
Specifically, our method includes a two-stage training phase to train a generative model incorporated with domain information and an inference phase to generate prompts directly.
In our method, the generation of soft prompts is exclusively handled by a generative model. Notably, the domain knowledge is stored within the generative model, making it possible for each image to generate an instance-specific prompt. Our method ensures the diversity of prompts and allows for the incorporation of domain-specific information.
We introduce our SPG method as follows.

\subsubsection{Training Stage I: Domain Prompt Labels Learning.}

To better adapt our SPG method to the DG problem, we introduce the concept of domain prompt labels. Each domain corresponds to a domain prompt label ${\mathbf{v}^{d_i}}$ where $d_i$ denotes $i$th domain, derived from training on the data of each domain with cross-entropy:

\begin{equation}\begin{aligned}\label{func:dpl}
{\mathbf{v}^{d_i}}^*=\arg \min _{\mathbf{v}} \mathbb{E}_{\mathbf{x}^{d_i}_j, {y}^{d_i}_j}\left[-\log p\left({y}_j^{d_i} \mid \mathbf{x}^{d_i}_j, \mathbf{v}^{d_i}\right)\right],
\end{aligned}\end{equation}
where $(\mathbf{x}^{d_i}_j$, ${y}^{d_i}_j)$ denotes images and labels of training samples in $i$th domain. The prompt design follows Zhou et al.~\cite{zhou2022learning} with text prompt tuning. These domain prompt labels represent optimal prompts for each domain, which encapsulate rich domain information.

\subsubsection{Training Stage II: Generative Model Pre-training.}

We adopt a simple yet efficient generative model, conditional generative adversarial net (CGAN)~\cite{mirza2014conditional}, to demonstrate the effectiveness of our method.
CGAN, an extension of the conventional generative adversarial network (GAN) framework, operates by conditioning the generation process on additional information, usually presented in the form of auxiliary input data or labels.
The CGAN architecture includes a generator $G$ and a discriminator $D$. The generator $G$ is guided by additional information, ensuring that the generated outputs align with the specified conditions. Meanwhile, the discriminator $D$ evaluates the authenticity of the real data and fake output, fostering a dynamic adversarial interplay that refines the overall generative capabilities of the model. In this work, the fake batch of images is randomly sampled from the dataset.

As shown in Figure~\ref{fig:stage2}, we adapt the CGAN model to serve the DG task of prompt generation, aligning the generated prompts with their corresponding images. 
Our objective is to learn the generator’s distribution $p_g$ over soft prompt $\mathbf{v}$, transferring the transferability from prompts to the generator. 
In the generator, we define a prior on input noise variables $\mathbf{z}$, and the joint hidden representation $[\mathbf{z}, f(\mathbf{x})]$ combines these noise variables with image embeddings $f(\mathbf{x})$ with concatenation operation.
In the discriminator, domain prompt labels and the generated prompt with image embeddings $f(\mathbf{x})$ serve as inputs to a discriminative function.
Therefore, the objective function $V(G, D)$ of a two-player minimax game can be formulated as:

\begin{equation}\begin{aligned}\label{func:adv_loss}
\min _G \max _D V(G, D)
&=\mathbb{E}_{\mathbf{v} \sim p_{\text {v}}(\mathbf{v})}[\log D(\mathbf{v} \mid f(\mathbf{x}))] \\
&\quad + \mathbb{E}_{{\mathbf{z} \sim p_{\text {z}}}(\mathbf{z})}[\log (1-D(G(\mathbf{z}|f(\mathbf{x})) ))].
\end{aligned}\end{equation}

The pre-trained CGAN aims to capture domain-invariant and domain-specific features, ensuring consistency across diverse domains. Additionally, its generator maintains prompt diversity, enhancing the model's adaptability and generalization capabilities to handle varied inputs.

\subsubsection{Inference.} The generator of CGAN is employed to produce a domain-specific soft prompt for each image of the target domain.
The probability of the image belonging to the $i$th class can be formulated as:

\begin{equation}\begin{aligned}\label{func:pro}
p(y=i \mid \mathbf{x})=\frac
{\exp (\langle w_i, f(\mathbf{x}) \rangle / \tau)}
{\sum^K_{j=1} \exp (\langle w_j, f(\mathbf{x}) \rangle / \tau)}, 
\end{aligned}\end{equation}

\begin{equation}\begin{aligned}\label{func:w}
\text{s.t.} ~~~~~~~~ w_i = g([G(\mathbf{z} \mid f(\mathbf{x})), c_i]),
\end{aligned}\end{equation}
where $\tau$ denotes temperature parameter, $K$ denotes the number of classes, $g$ denotes the text encoder, $G(\mathbf{z} \mid f(\mathbf{x}))$ denotes the domain-specific soft prompt, $c_i$ denotes $i$th tokenized class token, and $[\cdot,\cdot]$ denotes the concatenation operation.
In this way, exclusively relying on a generative model for soft prompt production provides benefits in terms of adaptability and dynamic prompt generation. Our method fosters diversity in prompt generation and enhances generalization across various tasks and domains.

\section{Experiments}
\label{sec:exp}

\subsection{Experimental Setting}

\subsubsection{Datasets.} 
Experiments are conducted on five popular benchmark datasets of DG, namely PACS~\cite{li2017deeper}, VLCS~\cite{fang2013unbiased}, OfficeHome~\cite{venkateswara2017deep}, TerraIncognita (TerraInc.)~\cite{beery2018recognition} and DomainNet~\cite{peng2019moment}.
The dataset details can be seen in the appendix.

\subsubsection{Baselines.} We consider two types of methods for comparisons. For the traditional DG methods, we report the performance of ERM~\cite{gulrajani2020search}, SWAD~\cite{cha2021swad}, MIRO~\cite{cha2022domain}. For CLIP-based methods, we compare SPG method with zero-shot CLIP (ZS-CLIP)~\cite{radford2021learning}, linear probing of CLIP (Lin. Prob.)~\cite{radford2021learning}, CoOp~\cite{zhou2022learning}, CoCoOp~\cite{zhou2022conditional}, VPT~\cite{jia2022visual}, VP~\cite{bahng2022exploring}, MaPLe~\cite{khattak2023maple}, DPL~\cite{zhang2023domain}.

\subsubsection{Experimental Setup.} 

ResNet50 (RN50)~\cite{he2016deep} and ViT-B/16 ~\cite{dosovitskiy2020image} are adopted as our backbones. 
For our SPG method, in the first stage, we train the domain prompt labels using the SGD optimizer with a batch size of 32 and the context length of the prompt label is 4. 
In the second stage, we train the CGAN model using the AdamW optimizer with weight decay 1e-4 and betas (0.9, 0.999), starting with a learning rate of 2e-3 for PACS, VLCS and TerraIncognita dataset, and a learning rate of 2e-4 for OfficeHome and DomainNet dataset. 
To enhance the stability of the CGAN, we incorporate a gradient clipping strategy to impose conditional constraints on the learnable parameter. 
For other CLIP-based methods, we set the learning rate initially to around 2e-3 and decay it using a cosine annealing rule with 20 epochs. We also set the batch size to 32. Moreover, the context tokens length is set to 2 for MaPLe, 10 for VPT and VP, and 16 for CoOp and CoCoOp.
We employ the training-domain validation set method for model selection for all methods, selecting the model that achieves the highest accuracy on the overall validation set.

\subsection{Comparison with SOTA Methods}

\subsubsection{Multi-source Domain Generalization.}

We follow the leave-one-domain-out evaluation protocol~\cite{gulrajani2020search} for multi-source domain generalization. In this protocol, the model excludes one domain from the training set in each evaluation round and is then tested on the excluded domain. This iterative process continues until each domain has been excluded once.

\begin{table*}[ht]
\centering
\caption{Comparisons with SOTA methods on five domain generalization benchmark datasets for multi-source DG in terms of mean leave-one-domain-out performance with ResNet50 and ViT-B/16 as the backbone (B.). The results marked by $\dag$ are the reported numbers from the original papers. Average accuracies and standard errors are reported from three trials. Bold denotes the best score.}
\label{tab:multi-DG}
\resizebox{\textwidth}{!}{
\begin{tabular}{
>{\centering\arraybackslash}m{3cm}|
>{\centering\arraybackslash}m{0.5cm}|
>{\centering\arraybackslash}m{2cm}
>{\centering\arraybackslash}m{1.5cm}
>{\centering\arraybackslash}m{2cm}
>{\centering\arraybackslash}m{1.7cm}
>{\centering\arraybackslash}m{1.7cm}
>{\centering\arraybackslash}m{1.2cm}}
\toprule
Method & B. & PACS & VLCS & OfficeHome & TerraInc. & DomainNet & Avg \\
\midrule
ERM$^\dag$~\cite{gulrajani2020search} & \multirow{3}{*}{\rotatebox{90}{RN50}} & 85.7\scriptsize{$\pm0.5$} & 77.4\scriptsize{$\pm0.3$} & 67.5\scriptsize{$\pm0.5$} & 47.2\scriptsize{$\pm0.4$} & 41.2\scriptsize{$\pm0.2$} & 63.8\scriptsize \\
MIRO$^\dag$~\cite{cha2022domain} &  & 85.4\scriptsize{$\pm0.4$} & 79.0\scriptsize{$\pm0.0$} & 70.5\scriptsize{$\pm0.4$} & 50.4\scriptsize{$\pm1.1$} & 44.3\scriptsize{$\pm0.2$} & 65.9 \\
SWAD$^\dag$~\cite{cha2021swad} &  & \textbf{88.1}\scriptsize{$\pm0.1$} & \textbf{79.1}\scriptsize{$\pm0.1$} & \textbf{70.6}\scriptsize{$\pm0.2$} & \textbf{50.0}\scriptsize{$\pm0.3$} & \textbf{46.5}\scriptsize{$\pm0.1$} & \textbf{66.9} \\
\midrule
ZS-CLIP~\cite{radford2021learning} & \multirow{7}{*}{\rotatebox{90}{CLIP RN50}} & 90.7\scriptsize{$\pm0.0$} & 80.0\scriptsize{$\pm0.0$} & 70.8\scriptsize{$\pm0.0$} & 23.8\scriptsize{$\pm0.0$} & 46.4\scriptsize{$\pm0.0$} & 62.3 \\
Lin. Prob.~\cite{radford2021learning} &  & 90.6\scriptsize{$\pm0.3$} & 79.8\scriptsize{$\pm0.4$} & 65.5\scriptsize{$\pm0.2$} & 33.0\scriptsize{$\pm1.2$} & 27.1\scriptsize{$\pm0.2$} & 59.2 \\
CoOp~\cite{zhou2022learning} & & 91.3\scriptsize{$\pm0.3$} & 81.4\scriptsize{$\pm0.2$} & 73.5\scriptsize{$\pm0.2$} & 33.2\scriptsize{$\pm3.4$} & 49.7\scriptsize{$\pm0.2$} & 65.9 \\
CoCoOp~\cite{zhou2022conditional} &  & 91.9\scriptsize{$\pm0.6$} & 81.8\scriptsize{$\pm0.3$} & 73.4\scriptsize{$\pm0.4$} & 34.1\scriptsize{$\pm3.0$} & 49.7\scriptsize{$\pm0.1$} & 66.3 \\
DPL~\cite{zhang2023domain} &  & 91.8\scriptsize{$\pm0.7$} & 80.8\scriptsize{$\pm0.8$} & 73.6\scriptsize{$\pm0.4$} & 34.4\scriptsize{$\pm1.0$} & 49.6\scriptsize{$\pm0.2$} & 66.0 \\
VP~\cite{bahng2022exploring} &  & 90.2\scriptsize{$\pm0.1$} & 80.5\scriptsize{$\pm0.3$} & 70.2\scriptsize{$\pm0.2$} & 25.6\scriptsize{$\pm1.0$} & 45.8\scriptsize{$\pm0.1$} & 62.4 \\
SPG (ours) &  & \textbf{92.8}\scriptsize{$\pm0.2$} & \textbf{84.0}\scriptsize{$\pm1.1$} & \textbf{73.8}\scriptsize{$\pm0.5$} & \textbf{37.5}\scriptsize{$\pm1.8$} & \textbf{50.1}\scriptsize{$\pm0.2$} & \textbf{67.5} \\
\midrule
ZS-CLIP~\cite{radford2021learning} & \multirow{9}{*}{\rotatebox{90}{CLIP ViT-B/16}} & 96.1\scriptsize{$\pm0.0$} & 82.3\scriptsize{$\pm0.0$} & 81.8\scriptsize{$\pm0.0$} & 33.8\scriptsize{$\pm0.0$} & 56.6\scriptsize{$\pm0.0$} & 70.2 \\
Lin. Prob.~\cite{radford2021learning} &  & 94.9\scriptsize{$\pm1.4$} & 77.5\scriptsize{$\pm0.7$} & 79.3\scriptsize{$\pm0.2$} & 44.6\scriptsize{$\pm2.1$} & 48.2\scriptsize{$\pm0.2$} & 68.9 \\
CoOp~\cite{zhou2022learning} &  & 96.4\scriptsize{$\pm0.3$} & 80.8\scriptsize{$\pm0.3$} & 83.0\scriptsize{$\pm0.1$} & 46.8\scriptsize{$\pm0.7$} & 59.5\scriptsize{$\pm0.2$} & 73.6 \\
CoCoOp~\cite{zhou2022conditional} &  & 96.7\scriptsize{$\pm0.2$} & 80.3\scriptsize{$\pm0.3$} & 83.4\scriptsize{$\pm0.2$} & 45.3\scriptsize{$\pm2.4$} & 59.4\scriptsize{$\pm0.2$} & 73.2 \\
DPL~\cite{zhang2023domain} &  & 96.4\scriptsize{$\pm0.3$} & 80.9\scriptsize{$\pm0.5$} & 83.0\scriptsize{$\pm0.3$} & 46.6\scriptsize{$\pm0.8$} & 59.5\scriptsize{$\pm0.3$} & 73.6 \\
VP~\cite{bahng2022exploring} &  & 95.8\scriptsize{$\pm0.1$} & 82.2\scriptsize{$\pm0.0$} & 81.2\scriptsize{$\pm0.2$} & 34.9\scriptsize{$\pm0.2$} & 56.5\scriptsize{$\pm0.0$} & 70.1 \\
VPT~\cite{jia2022visual} &  & 96.9\scriptsize{$\pm0.2$} & 82.0\scriptsize{$\pm0.2$} & 83.2\scriptsize{$\pm0.1$} & 46.7\scriptsize{$\pm0.6$} & 58.5\scriptsize{$\pm0.2$} & 73.6 \\
MaPLe~\cite{khattak2023maple} &  & 96.5\scriptsize{$\pm0.2$} & 82.2\scriptsize{$\pm0.2$} & 83.4\scriptsize{$\pm0.0$} & \textbf{50.2}\scriptsize{$\pm0.9$} & 59.5\scriptsize{$\pm0.3$} & 74.4 \\
SPG (ours) &  & \textbf{97.0}\scriptsize{$\pm0.5$} & \textbf{82.4}\scriptsize{$\pm0.4$} & \textbf{83.6}\scriptsize{$\pm0.4$} & \textbf{50.2}\scriptsize{$\pm1.2$} & \textbf{60.1}\scriptsize{$\pm0.5$} & \textbf{74.7} \\
\bottomrule
\end{tabular}
}
\end{table*}

In Table~\ref{tab:multi-DG}, we report the mean leave-one-domain-out performance of PACS, VLCS, Office-Home, TerraIncognita, and DomainNet for both ResNet50 and ViT-B/16 backbones. 
We can observe that SPG outperforms all other CLIP-based methods on five datasets with two backbones. 
For instance, with ResNet50 as the backbone, SPG can surpass zero-shot CLIP and the state-of-the-art (SOTA) CLIP-based method with a large margin of 13.7\% and 3.4\% for TerraIncognita. SPG also achieves a large improvement of around 4.0\% and 2.2\% compared with zero-shot CLIP and the SOTA method CoCoOp for VLCS. 
SPG achieves a convincing improvement of approximately 1.5\% on the averaged results of five benchmarks, establishing a new SOTA for the multi-source DG task. The results underscore the potential of generative models for prompt learning.

Meanwhile, SPG with ResNet50 as the backbone outperforms the SOTA traditional DG methods by a large margin of 4.7\%, 4.9\%, 3.2\%, and 3.6\% for PACS, VLCS, OfficeHome, and DomainNet datasets, respectively. 
We also observe that traditional DG methods achieve the SOTA performance on the TerraIncognita dataset, surpassing both the zero-shot CLIP model and our method. This discrepancy may arise from the fact that CLIP was not pre-trained on data similar to TerraIncognita, highlighting the need for further exploration of VLMs on unseen domain data and other downstream tasks.

\subsubsection{Single-source Domain Generalization.} The leave-all-but-one-domain-out evaluation protocol is adopted for single-source domain generalization. Under this protocol, all domains except one are included in the training set, and the model is then tested on the remaining domain.

\begin{table*}[ht]
\centering
\caption{Comparisons with CLIP-based fine-tuning methods on five domain generalization benchmark datasets for single-source DG in terms of leave-all-but-one-domain-out performance with ResNet50 as the backbone. Bold denotes the best scores.}
\label{tab:single-DG}
\resizebox{0.95\textwidth}{!}{
\begin{tabular}{
>{\centering\arraybackslash}m{3cm}|
>{\centering\arraybackslash}m{1.5cm}
>{\centering\arraybackslash}m{1.5cm}
>{\centering\arraybackslash}m{2cm}
>{\centering\arraybackslash}m{1.5cm}
>{\centering\arraybackslash}m{1.8cm}
>{\centering\arraybackslash}m{1.2cm}}
\toprule
Method & PACS & VLCS & OfficeHome & TerraInc. & DomainNet & Avg \\
\midrule
Lin. Prob.~\cite{radford2021learning} & 77.3 & 65.5 & 46.4 & 23.3 & 6.3 & 43.8 \\
CoOp~\cite{zhou2022learning} & 86.0 & 75.3 & 70.7 & 30.7 & 44.9 & 61.4 \\
CoCoOp~\cite{zhou2022conditional} & 88.1 & 68.2 & 70.6 & 25.6 & 45.5 & 59.6 \\
DPL~\cite{zhang2023domain} & 86.8 & 75.2 & 70.8 & 28.4 & 43.8 & 61.0 \\
SPG (ours) & \textbf{88.8} & \textbf{76.5} & \textbf{70.9} & \textbf{32.3} & \textbf{45.6} & \textbf{62.8}  \\
\bottomrule
\end{tabular}
}
\end{table*}

\begin{table*}[ht]
\centering
\caption{Comparisons with CLIP-based methods on four domain generalization benchmark datasets for cross-dataset generalization performance with ViT-B/16 as the backbone. Bold denotes the best scores.}
\label{tab:corss-DG}
\resizebox{0.83\textwidth}{!}{
\begin{tabular}{
>{\centering\arraybackslash}m{2.5cm}|
>{\centering\arraybackslash}m{1.5cm}
>{\centering\arraybackslash}m{1.5cm}
>{\centering\arraybackslash}m{2cm}
>{\centering\arraybackslash}m{2cm}
>{\centering\arraybackslash}m{1.2cm}}
\toprule
Method & PACS & VLCS & OfficeHome & TerraInc. & Avg \\
\midrule
ZS-CLIP~\cite{radford2021learning} & 94.5 & \textbf{80.5} & 82.1 & 30.9 & 72.0 \\
CoOp~\cite{zhou2022learning} & 96.2 & 72.8 & 83.1 & 32.1 & 71.1 \\
CoCoOp~\cite{zhou2022conditional} & 95.8 & 72.6 & 83.3 & 34.2 & 71.5 \\	 	
DPL~\cite{zhang2023domain} & 96.0 & 72.7 & 83.6 & 30.2 & 70.6 \\
VPT~\cite{jia2022visual} & 95.7 & 77.7 & 82.5 & 26.2 & 70.5 \\
MaPLe~\cite{khattak2023maple} & 95.1 & 74.4 & \textbf{83.9} & 27.3 & 70.2 \\
SPG (ours) & \textbf{96.7} & 76.7 & \textbf{83.9} & \textbf{38.0} & \textbf{73.8} \\
\bottomrule
\end{tabular}
}
\end{table*}

Table~\ref{tab:single-DG} shows the experimental results of the leave-all-but-one-domain-out performance of PACS, VLCS, Office-Home, TerraIncognita and DomainNet for ResNet50 backbone. SPG outperforms other CLIP-based fine-tuning methods on five datasets. We observe averaged improvements of 19.0\% and 1.4\% compared with linear probing of CLIP and the SOTA method CoOp, respectively. This demonstrates that by leveraging the generative model for prompt generation, SPG is able to produce more domain-relevant and adaptive prompts, leading to improved performance across different domains. 
The domain-wise results for both multi-source DG and single-source DG are provided in the appendix.

\subsubsection{Cross-Dataset Generalization.} We split the DomainNet subset into training and validation datasets. With training-domain validation set model selection, we train on training data of DomainNet and select the best model on validation data of DomainNet. Then we test the best model on four downstream datasets, i.e., PACS, VLCS, OfficeHome and TerraIncognita.

\begin{figure*}[ht]
  \centering
  \includegraphics[width=0.85\textwidth]{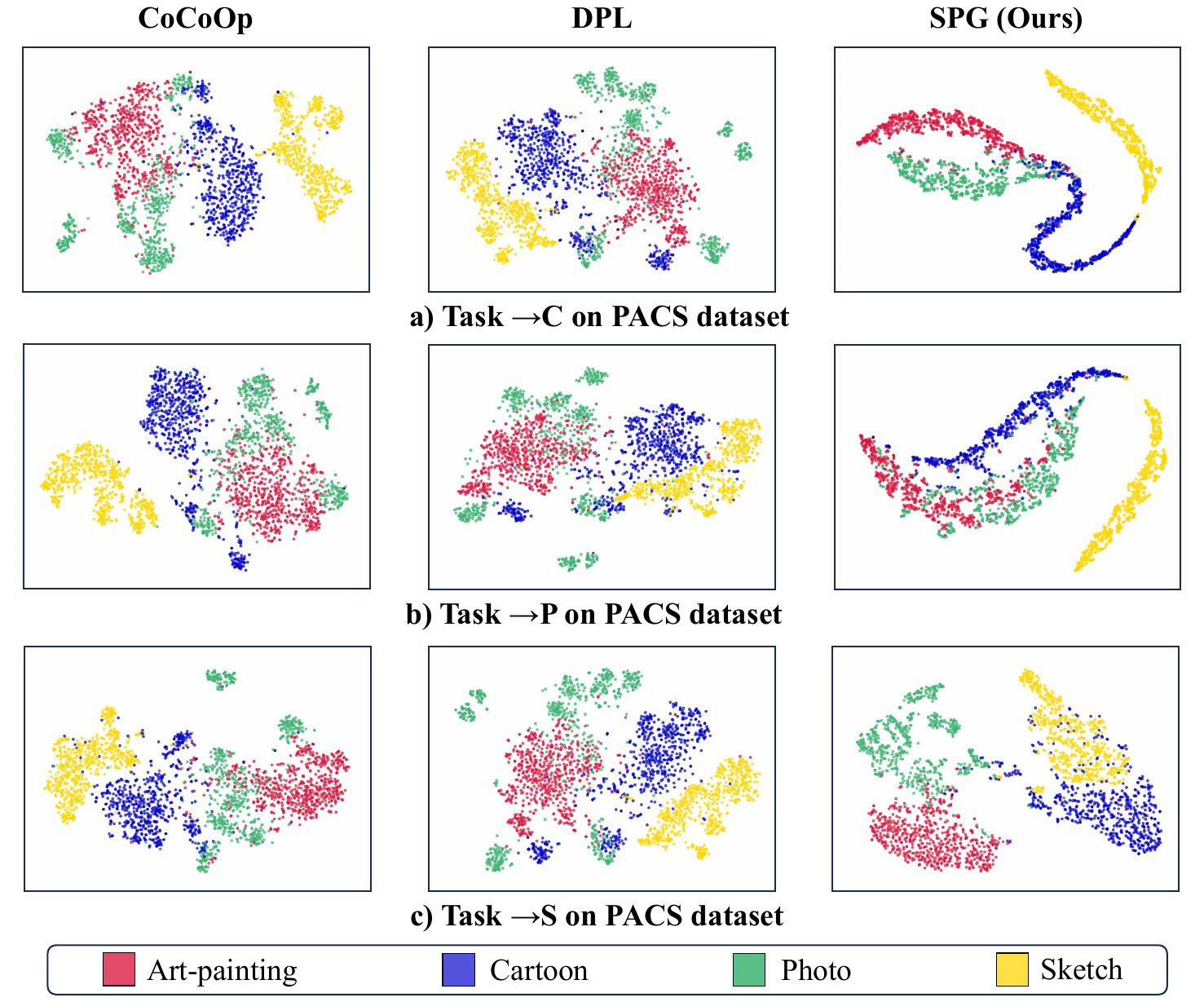}
  \caption{The t-sne visualization of the prompt embeddings for CoCoOp, DPL, and our SPG method. Multi-source domain generalization models on 3 tasks of PACS dataset are employed to obtain prompt embeddings. Different colors denote different classes.
  All the domains including the target domain are highly clustered in SPG.
  }
  \label{fig:tsne}
\end{figure*}

As shown in Table~\ref{tab:corss-DG}, SPG outperforms all other CLIP-based methods, indicating the effectiveness of our generative prompt learning method in tackling distribution shifts.
SPG achieves an average accuracy improvement of 1.8\% compared with the zero-shot CLIP.
It is also noteworthy that there is a stable improvement in the performance of baseline methods for the PACS and OfficeHome datasets, such as CoOp, CoCoOp, etc. 
We attribute this to the similarity in the distribution between DomainNet and these downstream datasets. However, when encountering a significant distribution shift, such as observed with the VLCS and TerraIncognita, the performance of these methods may decline due to the learned prompts not being able to generalize to the new distribution. 
For TerraIncognita dataset, our SPG method can mitigate this challenge to some extent, the generative model is capable of dynamically generating prompts for each image of unseen distribution, thus enabling more adaptable prompt generation.
Nevertheless, it remains particularly challenging for the VLCS dataset, which indicates a need for further exploration.

\subsection{Visualization}

\subsubsection{The t-SNE visualization.}

In Figure~\ref{fig:tsne}, we qualitatively
evaluate prompt embeddings synthesized by baseline methods and our SPG method for three multi-source domain generalization tasks of PACS using t-SNE visualization~\cite{donahue2014decaf}.
We aim to generate a domain-specific prompt for each image, which ensures that prompt diversity and domain knowledge benefit the DG tasks.
As illustrated in Figure~\ref{fig:tsne}, the prompts generated by our SPG method demonstrate a higher degree of clustering compared to prompts generated by alternative methods such as CoCoOp and DPL. This improved clustering indicates a higher degree of discriminative ability of the prompts, suggesting that our SPG method is more effective at capturing domain knowledge during prompt learning.
The clustered domain prompts may also facilitate a better understanding of the semantic relationship between images and text, allowing the model to focus more on comprehending class concepts when aligning with the image encoder.

\begin{figure*}[ht]
  \centering
  \includegraphics[width=0.95\textwidth]{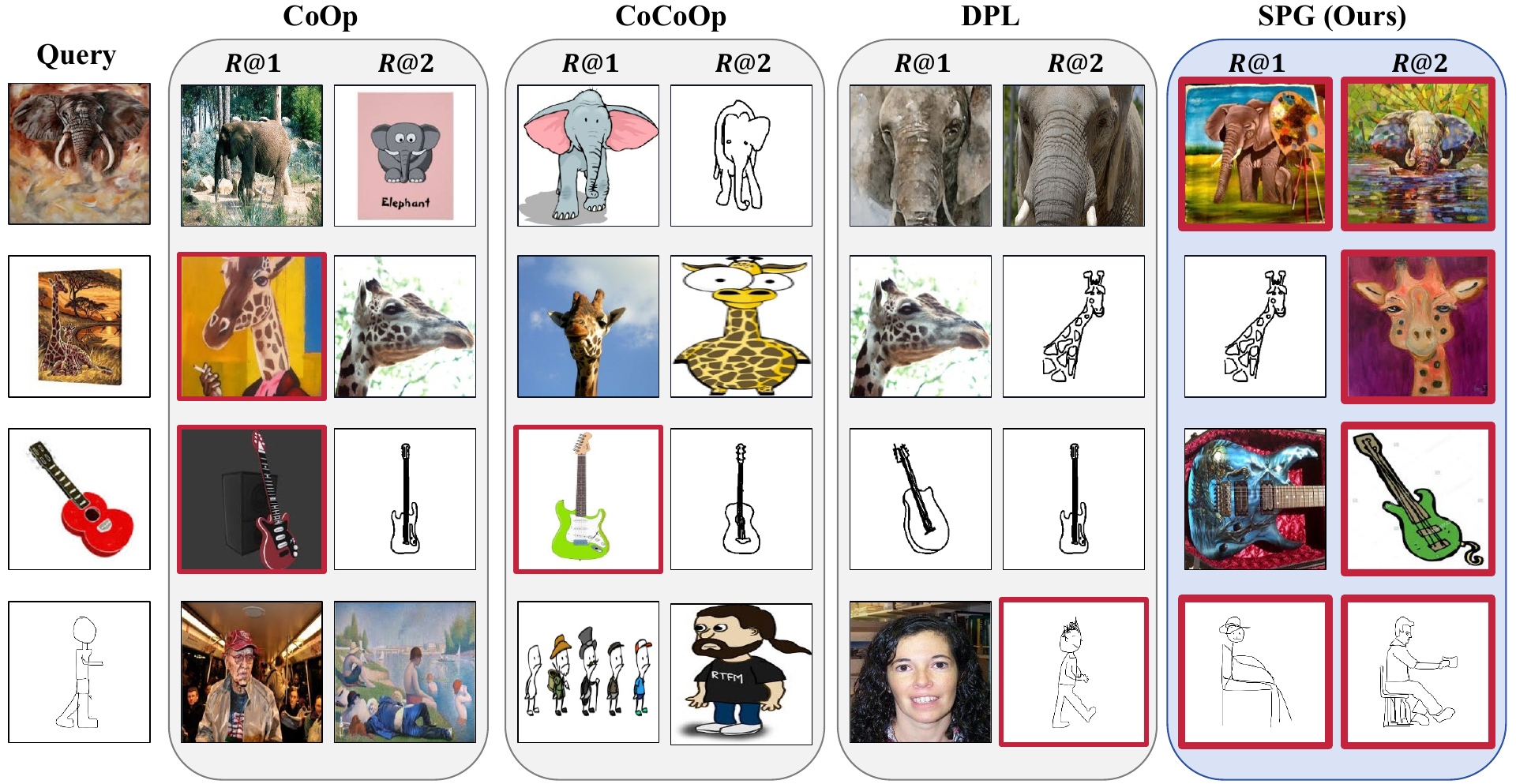}
  \caption{The image-prompt-image retrieval experiment is designed to demonstrate the correlation between the prompts and the images.  We present the top-2 results of image retrieval conducted using CoOp, CoCoOp, DPL, and our SPG method. 
  Images encased in red rectangles indicate instances where the query image and the retrieval image belong to the same domain.
  }
  \label{fig:image_re}
\end{figure*}

\subsubsection{Image retrieval.} We designed an image-prompt-image retrieval experiment to demonstrate the correlation between the generated prompt and the images in each domain. 
Specifically, we first sampled ${N}$-way $K$-shot images from each domain in the dataset and combined them into an image library, where ${N}$ is 7 and $K$ is 10 for PACS. 
Then, we randomly select one image as a query from the remaining images. The generated prompt of each query image is concatenated with the class token, which is passed through a text encoder to calculate the probability distribution with the features of each image in the image library.
Subsequently, we select the probability value corresponding to the class of the query image as the confidence score. 
These scores are then sorted in sequence, and the top-2 images are obtained as the results of the image retrieval.

In Figure~\ref{fig:image_re}, we can observe that the prompts generated by our SPG method are capable of retrieving images that are more closely aligned with the query image, implying that the prompts generated by our model for the query image are more closely related to domain-specific information. It shows the effectiveness of our SPG method in encoding domain-relevant information within the prompts with a generative model. 
While previous methods have not emphasized the importance of domain knowledge, our SPG method stands out by explicitly integrating domain-specific information into the prompt generation process. 

\subsection{Ablation Study}

\begin{table*}[ht]
\centering
\caption{Comparisons with different design of prompt on five domain generalization benchmark datasets for multi-source DG performance with ResNet50 as the backbone. O.H. denotes OfficeHome, and Do.Net denotes DomainNet. Bold denotes the best scores.}
\label{tab:abla_prompt}
\resizebox{\textwidth}{!}{
\begin{tabular}{
>{\centering\arraybackslash}m{3cm}
>{\centering\arraybackslash}m{1cm}
>{\centering\arraybackslash}m{1.6cm}
>{\centering\arraybackslash}m{1.6cm}
>{\centering\arraybackslash}m{1.8cm}
>{\centering\arraybackslash}m{1cm}
>{\centering\arraybackslash}m{1.2cm}
>{\centering\arraybackslash}m{1cm}
>{\centering\arraybackslash}m{1.3cm}
>{\centering\arraybackslash}m{1.3cm}}
\toprule
\multirow{2}{*}{Method} 
    & \multicolumn{4}{c}{Prompt Type} &  &  &  &  &  \\
    \cmidrule(r){2-5}
    & Fixed & Learnable & Conditional & Generative & PACS & VLCS & O.H. & TerraInc. & Do.Net \\
\midrule
Manual Prompt & $\checkmark$  &  &  &  & 90.7 & 80.0 & 70.8 & 23.8 & 46.4 \\
Mix-domain-prompt & $\checkmark$  & $\checkmark$  &  &  & 92.0 & 76.2 & 72.5 & 29.6 & 47.9  \\
All-domain-prompt & $\checkmark$  & $\checkmark$  &  &  & 91.3 & 81.4 & 73.2 & 33.2 & 49.7 \\
CoCoOp~\cite{zhou2022conditional} &  & $\checkmark$  & $\checkmark$  &  & 91.9 & 81.8 & 73.4 & 34.1 & 49.7 \\
DPL~\cite{zhang2023domain} &  & $\checkmark$  & $\checkmark$  &  & 91.8 & 80.8 & 73.6 & 34.4 & 49.6 \\
SPG (Ours) &  & $\checkmark$  &  & $\checkmark$  & \textbf{92.8} & \textbf{84.0} & \textbf{73.8} & \textbf{37.5} & \textbf{50.1} \\
\bottomrule
\end{tabular}
}
\end{table*}

\begin{table*}[ht]
\centering
\caption{Ablation on the context length of prompt on PACS for multi-DG performance with ResNet50 as the backbone.}
\label{tab:abla_token}
\resizebox{0.8\textwidth}{!}{
\begin{tabular}{
>{\centering\arraybackslash}m{3cm}
>{\centering\arraybackslash}m{1.5cm}
>{\centering\arraybackslash}m{1.5cm}
>{\centering\arraybackslash}m{1.5cm}
>{\centering\arraybackslash}m{1.5cm}
>{\centering\arraybackslash}m{1.5cm}}
\toprule
Context Length & Art & Cartoon & Photo & Sketch & Avg \\
\midrule
2 & 91.5 & \textbf{94.4} & 99.1 & 74.6 & 89.9 \\
4 & 92.8 & 93.8 & \textbf{99.5} & \textbf{85.1} & \textbf{92.8} \\
4 (random) & \textbf{93.0} & 93.4 & \textbf{99.5} & 82.9 & 92.2 \\
8 & 83.1 & 91.7 & 99.2 & 79.2 & 88.3 \\
16 & 84.7 & 93.6 & 98.3 & 82.0 & 89.7 \\
\bottomrule
\end{tabular}
}
\end{table*}

\begin{table*}[ht]
\centering
\caption{Ablation on the training samples of the generative model (CGAN) on PACS for multi-DG performance with ResNet50 as the backbone.}
\label{tab:abla_sample}
\resizebox{0.8\textwidth}{!}{
\begin{tabular}{
>{\centering\arraybackslash}m{3cm}
>{\centering\arraybackslash}m{1.5cm}
>{\centering\arraybackslash}m{1.5cm}
>{\centering\arraybackslash}m{1.5cm}
>{\centering\arraybackslash}m{1.5cm}
>{\centering\arraybackslash}m{1.5cm}}
\toprule
Training Samples & Art & Cartoon & Photo & Sketch & Avg \\
\midrule
20 & 30.1 & 34.9 & 23.4 & 34.5 & 30.7 \\
40 & 51.2 & 55.3 & 34.3 & 65.5 & 51.6 \\
60 & 73.8 & 76.7 & 56.8 & 81.1 & 72.1 \\
80 & 89.2 & 93.5 & 84.8 & \textbf{86.4} & 88.5 \\
full & \textbf{92.8} & \textbf{93.8} & \textbf{99.5} & 85.1 & \textbf{92.8} \\
\bottomrule
\end{tabular}
}
\end{table*}

\subsubsection{Ablation on the different design of prompt.} 
We have summarized four characteristics of prompts: fixed, learnable, conditional, and generative. Based on these characteristics, we selected five methods for comparison:
(1) Manual prompt: manually designed prompt as "a photo of a [CLS]", where [CLS] is the class token.
(2) Mix-domain-prompt: weighted sum of domain prompts obtained from each source domain. We set equal weights for each domain.
(3) All-domain-prompt: prompt obtained from all source domain data.
(4) CoCoOp. (5) DPL.

As shown in Table~\ref{tab:abla_prompt}, in the multi-source DG setting, the manual prompt performs the worst, highlighting the necessity of fine-tuning for downstream tasks.
For the learnable fixed prompt, the all-domain-prompt shows better performance than mix-domain-prompt. However, if the weights for mix-domain-prompt are more reasonable, there might be some improvement of performance, as some works have already attempted to explore this aspect~\cite{zheng2022prompt}.
Additionally, the conditional prompt methods outperform the fixed prompt methods, likely due to the integration of image information and a more diverse prompt. 
Finally, our generative prompt method performs the best, demonstrating its potential in prompt learning in VLMs.

\subsubsection{Sensitivity of context length of prompt.} 
As shown in Table~\ref{tab:abla_token}, we evaluate the effects of different context lengths of prompt for multi-source DG on PACS using the ResNet50 backbone. 
We mainly initialize the soft prompt in two ways. One is to initialize it with the embedding of the text "a photo of a", namely word embeddings-based initialization. The other one is to sample from a zero-mean Gaussian distribution with a standard deviation of 0.02 (marked as random). 
In Table~\ref{tab:abla_token}, we find that this initialization leads to slight improvement.
Overall, we observe that the prompt context length of 4 provides the optimal performance for our SPG method.

\subsubsection{Sensitivity to the number of training samples.} 
We sample different proportions of the training data for the generative model, including 20\%, 40\%, 60\%, 80\%, and all samples. 
Table~\ref{tab:abla_sample} presents the results of our multi-source domain generalization experiments on the PACS dataset, and we observe a consistent increase in accuracy for all four tasks on the PACS as the amount of data for training the generative model increases. 
The performance reached its optimum when using all the data.

\section{Conclusion}
\label{sec:conclu}

In this paper, we reframe the prompt learning framework from a generative perspective, and are the first to introduce the generative model into prompt learning in Vision Language Models (VLMs). We propose a simple yet efficient Soft Prompt Generation (SPG) method for the Domain Generalization (DG).
SPG is a new paradigm of prompt tuning, which consists of a two-stage training phase and an inference phase.
In the training phase, we introduce the concept of domain prompt labels, which are adopted to incorporate the generative model with domain knowledge.
During the inference phase, the generative model is employed to obtain domain-specific soft prompts for target domain data.
SPG relies exclusively on a generative model to directly produce soft prompts. This preserves the diversity of generated soft prompts, aiming to learn domain information with the generative model, including the target domain. 
Extensive experiments on three DG tasks of five DG benchmark datasets confirm the effectiveness of our proposed SPG method. Compared with traditional DG methods and CLIP-based approaches, SPG achieves new state-of-the-art performance for domain generalization.
We hope this research inspires future exploration into tapping the potential of generative models in prompt generation and learning.

\section*{Acknowledgments}
\label{sec:ack}

This work was supported by the National Natural Science Foundation of China under grant
number U21A20485, 62088102.

\bibliographystyle{splncs04}
\bibliography{main}

\clearpage

\appendix

This appendix is organized as follows: 

\begin{itemize}

\item Section~\ref{sec:DD} provides the detailed dataset information.

\item Section~\ref{sec:AD} provides the algorithm details.

\item Section~\ref{sec:ID} provides the additional training implementation details.

\item Section~\ref{sec:SCE} gives additional experiment results, including additional comparisons of domain-wise results of multi-source domain generalization and single-source domain generalization.

\end{itemize}

\section{Dataset Details}
\label{sec:DD}

\subsubsection{PACS~\cite{li2017deeper}} is a commonly used small-scaled dataset in the field of domain adaptation and domain generalization. It consists of 4 domains, a total of 9991 images, namely Photo (1,670 images), Art Painting (2,048 images), Cartoon (2,344 images), and Sketch (3,929 images). Each domain contains 7 categories.

\subsubsection{VLCS~\cite{fang2013unbiased}} is also a small-scaled benchmark dataset, a total of 7,510 images, including 4 domains: Caltech (991 images), LabelMe (1,859 images), Pascal (2,363 images) and Sun (2,297 images). Each domain contains 5 categories.

\subsubsection{Office-Home~\cite{venkateswara2017deep}} is a medium-scaled benchmark for domain adaptation and domain generalization. It contains a total of around 15,500 images from 4 distinct domains: Art (2,427 images), Clip Art (4,365 images), Product (4,439 images), and Real World (4,357 images). Each domain contains objects from 65 categories commonly found in office and home environments.

\subsubsection{TerraIncognita~\cite{beery2018recognition}} is a large-scaled benchmark for visual recognition. It contains 243,187 images from 140 camera locations. For DG, a subset is selected that includes 4 domains: Location38 (9,736 images), Location43 (3,970 images), Location46 (5,883 images) and Location100 (4,741 images). Each domain contains animals from 10 categories found in the wild.

\subsubsection{DomainNet~\cite{peng2019moment}} is a large-scaled benchmark for domain adaptation and domain generalization.  It contains a total of around 586,575 images from 6 distinct domains: Clipart (48,129 images), Infograph (51,605 images), Painting (72,266 images), Quickdraw (172,500 images), Real (172,947 images), Sketch (69,128 images). Each domain includes 345 categories of objects. We sample around 20 thousand data from all domains as the subset.

\section{Algorithm Details}
\label{sec:AD}

The overall framework of the pseudo-code of SPG is described in Algorithm~\ref{alg:spg_stage1} and Algorithm~\ref{alg:spg_stage2}.
Algorithm~\ref{alg:spg_stage1} demonstrates the process of stage I: Domain Prompt Labels Learning, and Algorithm~\ref{alg:spg_stage2} shows the process of stage II: Generative Model Pre-training.

\begin{algorithm}
    \caption{\mbox{Soft Prompt Generation - Domain Prompt Labels Learning}}
    \label{alg:spg_stage1}
    \textbf{Requirement:} pre-defined ${N}_c$ class names in the target task \\
    \textbf{Input:} images and labels of training samples $(\mathbf{x}^{{d}_{i}}_{j}, {y}^{{d}_{i}}_{j})$, number of training iterations $L$ \\
    \textbf{Output:} ${N}_{d}$ domain prompt labels
    \begin{algorithmic}[1]
        \Statex \commentcolor{\# learn prompt label on each domain separately}
        \For {$i=1,2,\ldots,{N}_{d}$} 
            \Statex \commentcolor{\;\;\;\;\;\,\# initialize $i$-th domain prompt label with prompt prefix $\mathbf{v}^p$ and learnable vector $\mathbf{v}^i$.}
            \State $\mathbf{v}^{{d}_{i}} \leftarrow \mathtt{initialize}(\mathbf{v}^p, \mathbf{v}^i)$
        
            \Statex \commentcolor{\;\;\;\;\;\,\# $L$ training iterations for learning each domain prompt label}
            \For {$\mathrm{iteration}=1,2,\ldots,L$} 
                \Statex \commentcolor{\;\;\;\;\;\;\;\;\;\;\,\,\# update learnable vector ${v}^{d_i}$}
                \State ${\mathbf {v}^{d_i}}^*=\arg \min _{\mathbf {v}} \mathbb {E}_{\mathbf {x}^{d_i}_j, {y}^{d_i}_j} [-\log p ({y}_j^{d_i} \mid \mathbf {x}^{d_i}_j, \mathbf {v}^{d_i} ) ]$
                \State Update $\mathbf{v}^{d_i}$ by gradient descent
            \EndFor
        \EndFor
        \State Store optimal domain prompt label $\mathbf{v}^{d_i}$ for each domain
    \end{algorithmic} 
\end{algorithm} 

\begin{algorithm}
    \caption{\mbox{Soft Prompt Generation - Generative Model Pre-training}}
    \label{alg:spg_stage2}
    \textbf{Requirement:} A CGAN with a generator $G$ and a discriminator $D$, real vector $\mathbf{v}_{\mathrm{real}}$ and fake vector $\mathbf{v}_{\mathrm{fake}}$, and domain prompt labels $\mathbf{v}^{d_i}$ \\
    \textbf{Input:} image embeddings $f(\mathbf{x})$, number of training iterations $L$ \\
    \textbf{Output:} optimal prompt for each image
    \begin{algorithmic}[1]
        \Statex \commentcolor{\# $L$ training iterations}
        \For {$\mathrm{iteration}=1,2,\ldots,L$} 
            \Statex \commentcolor{\;\;\;\;\;\,\,\# define input noise variable $\mathbf{z}$ and combines these noise variables with image embeddings $f(\mathbf{x})$ as input of generator G and output $\mathbf{v}_g$}
            \State $\mathbf{v}_g=G(input) \leftarrow input=\mathtt{concat}([ \mathbf{z}, f(\mathbf{x}) ])$
            \Statex \commentcolor{\;\;\;\;\;\,\,\# discriminator determines the authenticity of domain prompt labels $\mathbf{v}^{d_i}$ and generated prompt $\mathbf{v}_g$}
            \State $\mathbf{v}_{d, \mathrm{real}} = D(\mathbf{v}^{d_i}) $, $\mathbf{v}_{d, \mathrm{fake}} = D(\mathbf{v}_g) $
            
            \Statex \commentcolor{\;\;\;\;\;\,\,\# compute $\mathcal{L}_{\mathrm{real}}$ with discriminator\_real output $\mathbf{v}_{d, \mathrm{real}}$ and pre-defined real vector $\mathbf{v}_{\mathrm{real}}$}
            \State $\mathcal{L}_{\mathrm{real}} \leftarrow \mathtt{mse\_loss}(\mathbf{v}_{d, \mathrm{real}}, \mathbf{v}_{\mathrm{real}})$
            \Statex \commentcolor{\;\;\;\;\;\,\,\# compute $\mathcal{L}_{\mathrm{fake}}$ with discriminator\_fake output $\mathbf{v}_{d, \mathrm{fake}}$ and pre-defined fake vector $\mathbf{v}_{\mathrm{fake}}$}
            \State $\mathcal{L}_{\mathrm{fake}} \leftarrow \mathtt{mse\_loss}(\mathbf{v}_{d, \mathrm{fake}}, \mathbf{v}_{\mathrm{fake}})$
            \State $\mathcal{L}_{\mathrm{discriminator}} \leftarrow \mathcal{L}_{\mathrm{real}} + \mathcal{L}_{\mathrm{fake}}$ 
            \State Update parameters of $D$ using $\mathcal{L}_{\mathrm{discriminator}}$ by gradient descent
            \Statex \commentcolor{\;\;\;\;\;\,\,\# compute $\mathcal{L}_{\mathrm{generator}}$ with discriminator\_fake output $\mathbf{v}_{d, \mathrm{fake}}$ and pre-defined real vector $\mathbf{v}_{\mathrm{real}}$}
            \State $\mathcal{L}_{\mathrm{generator}} \leftarrow \mathtt{mse\_loss}(\mathbf{v}_{d, \mathrm{fake}}, \mathbf{v}_{\mathrm{real}})$
            \State Update parameters of $G$ using $\mathcal{L}_{\mathrm{generator}}$ by gradient descent
        \EndFor
        \State Generate the optimal prompt for each input image 
    \end{algorithmic} 
\end{algorithm}

\section{Implementation Details}
\label{sec:ID}

For our proposed SPG, in the first phase of the training stage, we employ the text prompt~\cite{zhou2022learning} as our prompt design, which is also the prototype for the domain prompt labels and generated prompts. We initialize the context with the phrase "a photo of a" and set the prompt's context length to 4.
In the second phase of the training stage, we train the CGAN model on different domains of various datasets, employing a tailored set of training parameters. 
Specifically, we set the batch size to around 32 and adjust the initial learning rate between 1e-4 and 2e-3 on different datasets, 
We employ a cosine learning rate scheduler and conduct training for 70 to 100 epochs, incorporating a linear learning rate warm-up phase at 1e-5 over the first 4 epochs.
For optimization, we use the AdamW optimizer, configuring it with a weight decay of 1e-4 and beta values set to (0.9, 0.999).

Meanwhile, given the observed instability in CGAN training, we implement the gradient clipping strategy to control the magnitude of gradients within the generator and discriminator networks.
Specifically, for the discriminator, we establish norm upper limits for general weights and biases in the range of 5e-2 to 5e-1, while setting those for particular weights and biases at a ceiling of 5.
For the generator, we set the norm upper limit for universal weights between 5e-3 and 5e-2, the norm upper limit for universal biases between 5e-8 and 5e-7, and the norm upper limit for special biases between 0.5 and 5.
This strategy aims to enhance training stability by preventing excessive gradient values.

\section{Supplement Experiments}
\label{sec:SCE}

The supplement experiments mainly demonstrate the domain-wise results of multi-source DG and single-source DG.

\subsection{Multi-source DG Comparisons}

Table~\ref{tab:multi-DG_pacs}$\sim$\ref{tab:multi-DG_domainnet} demonstrate the per-domain multi-source DG top-1 classification accuracy on PACS, VLCS, OfficeHome, TerraIncognita, and DomainNet, respectively.
We observe that although SPG may not achieve the best performance in every individual domain, it consistently reaches state-of-the-art levels across five datasets under two different backbones on average. 
For some tasks, such as the target domain is Sketch in PACS and Sun in VLCS, SPG outperforms the previous SOTA method with a large margin of 4.3\% and 3.6\%, respectively.

\subsection{Single-source DG Comparisons}

Table~\ref{tab:single-DG_pacs}$\sim$\ref{tab:single-DG_domainnet} demonstrate the per-domain single-source DG top-1 classification accuracy on PACS, VLCS, OfficeHome, TerraIncognita, and DomainNet, respectively.
Relying on a single domain for generalization can significantly degrade performance. 
While our approach may not be optimal in certain cases, it still achieves state-of-the-art performance on average.

\begin{table*}[ht]
\centering
\caption{Comparisons with SOTA methods on PACS for multi-source DG in terms of mean leave-one-domain-out performance with ResNet50 and ViT-B/16 as the backbone (B.). Average accuracies are reported from three trials. Bold denotes the best scores.}
\label{tab:multi-DG_pacs}
\resizebox{0.8\textwidth}{!}{
\begin{tabular}{
>{\centering\arraybackslash}m{2.5cm}|
>{\centering\arraybackslash}m{0.5cm}|
>{\centering\arraybackslash}m{1.5cm}
>{\centering\arraybackslash}m{1.5cm}
>{\centering\arraybackslash}m{1.5cm}
>{\centering\arraybackslash}m{1.5cm}
>{\centering\arraybackslash}m{1.5cm}}
\toprule
Method & B. & Art & Cartoon & Photo & Sketch & Avg \\
\midrule
ZS-CLIP~\cite{radford2021learning} & \multirow{7}{*}{\rotatebox{90}{RN50}} & 90.9 & 93.3 & 99.2 & 79.5 & 90.7 \\
Lin. Prob.~\cite{radford2021learning} &  & 90.8 & 92.7 & 99.1 & 79.8 & 90.6 \\
CoOp~\cite{zhou2022learning} &  & 92.0 & 93.8 & 98.6 & 80.7 & 91.3 \\
CoCoOp~\cite{zhou2022conditional} &  & 93.1 & \textbf{94.3} & 99.3 & 80.8 & 91.9 \\
DPL~\cite{zhang2023domain} &  & \textbf{93.6} & 93.8 & 99.0 & 80.7 & 91.8 \\
VP~\cite{bahng2022exploring} &  & 90.6 & 92.7 & 99.3 & 78.0 & 90.2 \\
SPG (ours) &  & 92.8 & 93.8 & \textbf{99.5} & \textbf{85.1} & \textbf{92.8} \\
\midrule
ZS-CLIP~\cite{radford2021learning} & \multirow{9}{*}{\rotatebox{90}{ViT-B/16}} & 97.2 & \textbf{99.1} & \textbf{99.9} & 88.2 & 96.1 \\
Lin. Prob.~\cite{radford2021learning} &  & 96.2 & 94.7 & 98.7 & 90.1 & 94.9 \\
CoOp~\cite{zhou2022learning} &  & 97.7 & 98.4 & 99.6 & 90.0 & 96.4 \\
CoCoOp~\cite{zhou2022conditional} &  & 97.7 & {99.0} & 99.8 & 90.4 & 96.7 \\
DPL~\cite{zhang2023domain} &  & 97.8 & 98.5 & \textbf{99.9} & 89.5 & 96.4 \\
VP~\cite{bahng2022exploring} &  & 96.9 & 98.9 & \textbf{99.9} & 87.3 & 95.8 \\
VPT~\cite{jia2022visual} &  & \textbf{97.9} & 98.9 &\textbf{99.9} & 91.0 & 96.9 \\
MaPLe~\cite{khattak2023maple} &  & \textbf{97.9} & 98.7 & 99.7 & 89.8 & 96.5 \\
SPG (ours) &  & 97.7 & 99.0 & \textbf{99.9} & \textbf{91.3} & \textbf{97.0} \\
\bottomrule
\end{tabular}
}
\end{table*}

\begin{table*}[ht]
\centering
\caption{Comparisons with SOTA methods on VLCS for multi-source DG in terms of mean leave-one-domain-out performance with ResNet50 and ViT-B/16 as the backbone (B.). Average accuracies are reported from three trials. Bold denotes the best scores.}
\label{tab:multi-DG_vlcs}
\resizebox{0.8\textwidth}{!}{
\begin{tabular}{
>{\centering\arraybackslash}m{2.5cm}|
>{\centering\arraybackslash}m{0.5cm}|
>{\centering\arraybackslash}m{1.5cm}
>{\centering\arraybackslash}m{1.5cm}
>{\centering\arraybackslash}m{1.5cm}
>{\centering\arraybackslash}m{1.5cm}
>{\centering\arraybackslash}m{1.5cm}}
\toprule			
Method & B. & Caltech & LableMe & Pascal & Sun & Avg \\
\midrule
ZS-CLIP~\cite{radford2021learning} & \multirow{7}{*}{\rotatebox{90}{RN50}} & 99.4 & 64.9 & 84.1 & 71.6 & 80.0 \\
Lin. Prob.~\cite{radford2021learning} &  & 99.3 & 61.1 & 81.8 & 76.9 & 79.8 \\
CoOp~\cite{zhou2022learning} &  & 99.7 & 64.0 & 84.7 & 77.3 & 81.4 \\
CoCoOp~\cite{zhou2022conditional} &  & 99.7 & 63.7 & 84.8 & 78.8 & 81.8 \\
DPL~\cite{zhang2023domain} &  & \textbf{99.8} & 62.5 & 84.5 & 76.3 & 80.8 \\
VP~\cite{bahng2022exploring} &  & 99.6 & 66.3 & 84.6 & 71.5 & 80.5 \\
SPG (ours) &  & 99.5 & \textbf{68.7} & \textbf{85.4} & \textbf{82.4} & \textbf{84.0} \\
\midrule
zero-shot CLIP & \multirow{9}{*}{\rotatebox{90}{ViT-B/16}} & 99.9 & {68.6} & 85.9 & 74.8 & 82.3 \\
Lin. Prob.~\cite{radford2021learning} &  & 95.9 & 63.7 & 76.3 & 74.2 & 77.5 \\
CoOp~\cite{zhou2022learning} &  & 99.6 & 61.4 & 84.6 & 77.5 & 80.8 \\
CoCoOp~\cite{zhou2022conditional} &  & 99.9 & 59.7 & 85.9 & 75.5 & 80.3 \\
DPL~\cite{zhang2023domain} &  & 99.8 & 61.5 & 84.6 & 77.8 & 80.9 \\
VP~\cite{bahng2022exploring} &  & \textbf{100.0} & \textbf{68.5} & \textbf{86.2} & 73.9 & 82.2 \\
VPT~\cite{jia2022visual} &  & 99.9 & 64.8 & 85.2 & 78.2 & 82.0 \\
MaPLe~\cite{khattak2023maple} &  & 98.3 & 64.8 & 85.1 & 80.6 & 82.2 \\
SPG (ours) &  & 99.7 & 64.7 & 84.4 & \textbf{80.7} & \textbf{82.4} \\
\bottomrule
\end{tabular}
}
\end{table*}

\begin{table*}[ht]
\centering
\caption{Comparisons with SOTA methods on OfficeHome for multi-source DG in terms of mean leave-one-domain-out performance with ResNet50 and ViT-B/16 as the backbone (B.). Average accuracies are reported from three trials. Bold denotes the best scores.}
\label{tab:multi-DG_officehome}
\resizebox{0.8\textwidth}{!}{
\begin{tabular}{
>{\centering\arraybackslash}m{2.5cm}|
>{\centering\arraybackslash}m{0.5cm}|
>{\centering\arraybackslash}m{1.5cm}
>{\centering\arraybackslash}m{1.5cm}
>{\centering\arraybackslash}m{1.5cm}
>{\centering\arraybackslash}m{1.5cm}
>{\centering\arraybackslash}m{1.5cm}}
\toprule
Method & B. & Art & Clipart & Product & Real & Avg \\
\midrule
ZS-CLIP~\cite{radford2021learning} & \multirow{7}{*}{\rotatebox{90}{RN50}} & 69.0 & 53.5 & 80.1 & 80.5 & 70.8 \\
Lin. Prob.~\cite{radford2021learning} &  & 62.0 & 49.0 & 73.6 & 77.4 & 65.5 \\
CoOp~\cite{zhou2022learning} &  & 71.3 & 56.1 & 83.2 & 83.2 & 73.5 \\
CoCoOp~\cite{zhou2022conditional} &  & 70.3 & \textbf{56.7} & 83.4 & 83.3 & 73.4 \\
DPL~\cite{zhang2023domain} &  & \textbf{71.5} & 56.2 & 83.5 & 83.1 & 73.6 \\
VP~\cite{bahng2022exploring} &  & 67.7 & 52.5 & 80.0 & 80.4 & 70.2 \\
SPG (ours) &  & 71.3 & 55.6 & \textbf{84.8} & \textbf{83.4} & \textbf{73.8} \\
\midrule
ZS-CLIP~\cite{radford2021learning} & \multirow{9}{*}{\rotatebox{90}{ViT-B/16}} & 80.1 & 70.0 & 88.2 & 89.0 & 81.8 \\
Lin. Prob.~\cite{radford2021learning} &  & 73.5 & 69.9 & 87.4 & 86.4 & 79.3 \\
CoOp~\cite{zhou2022learning} &  & 81.2 & 72.0 & 89.7 & 89.2 & 83.0 \\
CoCoOp~\cite{zhou2022conditional} &  & 81.8 & 71.7 & \textbf{90.3} & 89.7 & 83.4 \\
DPL~\cite{zhang2023domain} &  & 81.0 & 71.2 & 90.0 & 89.6 & 83.0 \\
VP~\cite{bahng2022exploring} &  & 79.8 & 69.1 & 87.4 & 88.6 & 81.2 \\
VPT~\cite{jia2022visual} &  & 80.9 & 72.5 & 89.0 & 90.4 & 83.2 \\
MaPLe~\cite{khattak2023maple} &  & \textbf{81.6} & 72.6 & 90.2 & 89.0 & 83.4 \\
SPG (ours) &  & \textbf{81.6} & \textbf{72.7} & 90.2 & \textbf{89.9} & \textbf{83.6} \\
\bottomrule
\end{tabular}
}
\end{table*}

\begin{table*}[ht]
\centering
\caption{Comparisons with SOTA methods on TerraIncognita for multi-source DG in terms of mean leave-one-domain-out performance with ResNet50 and ViT-B/16 as the backbone (B.). Average accuracies are reported from three trials. Bold denotes the best scores.}
\label{tab:multi-DG_terra}
\resizebox{0.95\textwidth}{!}{
\begin{tabular}{
>{\centering\arraybackslash}m{2.5cm}|
>{\centering\arraybackslash}m{0.5cm}|
>{\centering\arraybackslash}m{2cm}
>{\centering\arraybackslash}m{2cm}
>{\centering\arraybackslash}m{2cm}
>{\centering\arraybackslash}m{2cm}
>{\centering\arraybackslash}m{1.5cm}}
\toprule
Method & B. & Location38 & Location43 & Location46 & Location100 & Avg \\
\midrule
ZS-CLIP~\cite{radford2021learning} & \multirow{7}{*}{\rotatebox{90}{RN50}} & 28.4 & 32.8 & 24.0 & 10.1 & 23.8 \\
Lin. Prob.~\cite{radford2021learning} &  & 33.0 & 42.7 & 31.9 & 24.4 & 33.0 \\
CoOp~\cite{zhou2022learning} &  & 25.6 & \textbf{43.5} & \textbf{34.5} & 29.2 & 33.2 \\
CoCoOp~\cite{zhou2022conditional} &  & 35.9 & 42.1 & 32.5 & 25.8 & 34.1 \\
DPL~\cite{zhang2023domain} &  & 36.0 & 41.1 & 32.9 & 27.6 & 34.4 \\
VP~\cite{bahng2022exploring} &  & 28.8 & 34.0 & 26.8 & 12.6 & 25.6 \\
SPG (ours) &  & \textbf{42.1} & 38.9 & 32.1 & \textbf{36.8} & \textbf{37.5} \\
\midrule
ZS-CLIP~\cite{radford2021learning} & \multirow{9}{*}{\rotatebox{90}{ViT-B/16}} & 20.5 & 32.8 & 29.6 & \textbf{52.4} & 33.8 \\
Lin. Prob.~\cite{radford2021learning} &  & 48.0 & 50.5 & 43.8 & 44.0 & 46.6 \\
CoOp~\cite{zhou2022learning} &  & \textbf{53.3} & 47.4 & 41.1 & 45.5 & 46.8 \\
CoCoOp~\cite{zhou2022conditional} &  & 51.6 & 46.9 & 39.3 & 43.2 & 45.3 \\
DPL~\cite{zhang2023domain} &  & 54.3 & 49.0 & 41.6 & 41.6 & 46.6 \\
VP~\cite{bahng2022exploring} &  & 20.2 & 34.3 & 32.8 & 52.3 & 34.9 \\
VPT~\cite{jia2022visual} &  & 46.8 & 52.8 & 41.8 & 45.5 & 46.7 \\
MaPLe~\cite{khattak2023maple} &  & 52.4 & \textbf{53.0} & 43.1 & \textbf{52.4} & \textbf{50.2} \\
SPG (ours) &  & 51.0 & 49.2 & \textbf{50.7} & 49.8 & \textbf{50.2}\\
\bottomrule
\end{tabular}
}
\end{table*}

\begin{table*}[ht]
\centering
\caption{Comparisons with SOTA methods on DomainNet for multi-source DG in terms of mean leave-one-domain-out performance with ResNet50 and ViT-B/16 as the backbone (B.). Average accuracies are reported from three trials. Bold denotes the best scores.}
\label{tab:multi-DG_domainnet}
\resizebox{\textwidth}{!}{
\begin{tabular}{
>{\centering\arraybackslash}m{2.5cm}|
>{\centering\arraybackslash}m{0.5cm}|
>{\centering\arraybackslash}m{1.5cm}
>{\centering\arraybackslash}m{1.5cm}
>{\centering\arraybackslash}m{1.8cm}
>{\centering\arraybackslash}m{1.5cm}
>{\centering\arraybackslash}m{1.5cm}
>{\centering\arraybackslash}m{1.5cm}
>{\centering\arraybackslash}m{1.5cm}}
\toprule
Method & B. & Clipart & Infograph & Painting & Quickdraw & Real & Sketch & Avg \\
\midrule
ZS-CLIP~\cite{radford2021learning} & \multirow{7}{*}{\rotatebox{90}{RN50}} & 52.7 & 40.5 & 53.2 & 5.7 & 77.1 & 49.3 & 46.4 \\
Lin. Prob.~\cite{radford2021learning} &  & 34.6 & 24.7 & 35.3 & 4.1 & 28.2 & 35.9 & 27.1 \\
CoOp~\cite{zhou2022learning} &  & 57.0 & 43.9 & 58.1 & 7.8 & 78.8 & 52.6 & 49.7 \\
CoCoOp~\cite{zhou2022conditional} &  & 57.0 & \textbf{44.0} & 58.3 & 7.8 & 78.9 & 52.0 & 49.7 \\
DPL~\cite{zhang2023domain} &  & 56.7 & 43.9 & 57.9 & \textbf{7.9} & 78.2 & 53.0 & 49.6 \\
VP~\cite{bahng2022exploring} &  & 52.4 & 40.3 & 52.7 & 5.3 & 76.8 & 47.1 & 45.8 \\
SPG (ours) &  & \textbf{57.3} & 41.7 & \textbf{58.3} & \textbf{7.9} & \textbf{80.0} & \textbf{55.5} & \textbf{50.1} \\
\midrule
ZS-CLIP~\cite{radford2021learning} & \multirow{9}{*}{\rotatebox{90}{ViT-B/16}} & 70.2 & 46.3 & 65.0 & 13.0 & 83.0 & 62.0 & 56.6 \\
Lin. Prob.~\cite{radford2021learning} &  & 62.9 & 35.4 & 56.8 & 11.3 & 65.8 & 56.7 & 48.2 \\
CoOp~\cite{zhou2022learning} &  & 72.7 & 50.2 & 68.5 & 15.6 & 84.2 & 65.9 & 59.5 \\
CoCoOp~\cite{zhou2022conditional} &  & 72.1 & 50.4 & 67.9 & 15.8 & \textbf{84.4} & 65.5 & 59.4 \\
DPL~\cite{zhang2023domain} &  & 72.5 & 50.4 & 68.3 & 15.8 & 83.9 & 66.0 & 59.5 \\
VP~\cite{bahng2022exploring} &  & 70.1 & 45.5 & 64.6 & 14.1 & 82.7 & 62.0 & 56.5 \\
VPT~\cite{jia2022visual} &  & 71.0 & 48.5 & 66.2 & 16.3 & 83.6 & 65.2 & 58.5 \\
MaPLe~\cite{khattak2023maple} &  & \textbf{73.1} & 49.9 & 67.8 & \textbf{16.6} & 83.5 & 65.9 & 59.5 \\
SPG (ours) &  & 68.7 & \textbf{50.2} & \textbf{73.2} & \textbf{16.6} & 83.3 & \textbf{68.5} & \textbf{60.1} \\
\bottomrule
\end{tabular}
}
\end{table*}

\begin{table*}[ht]
\centering
\caption{Comparisons with CLIP-base fine-tuning methods on PACS for single-source DG in terms of leave-all-but-one-domain-out performance with ResNet50 as the backbone. Bold denotes the best scores.}
\label{tab:single-DG_pacs}
\resizebox{0.85\textwidth}{!}{
\begin{tabular}{
>{\centering\arraybackslash}m{3cm}|
>{\centering\arraybackslash}m{1.5cm}
>{\centering\arraybackslash}m{1.5cm}
>{\centering\arraybackslash}m{1.5cm}
>{\centering\arraybackslash}m{1.5cm}
>{\centering\arraybackslash}m{1.5cm}
>{\centering\arraybackslash}m{1.2cm}}
\toprule
Method & Art & Cartoon & Photo & Sketch & Avg \\
\midrule    
Lin. Prob.~\cite{radford2021learning} & 79.2 & 82.2 & 76.7 & 71.2 & 77.3 \\
CoOp~\cite{zhou2022learning} & \textbf{91.4} & 84.5 & 82.5 & 85.4 & 86.0 \\
CoCoOp~\cite{zhou2022conditional} & 91.1 & 85.7 & 88.2 & 87.5 & 88.1 \\
DPL~\cite{zhang2023domain} & 90.4 & 83.7 & 84.5 & 88.5 & 86.8 \\
SPG (ours) & 90.5 & \textbf{87.4} & \textbf{88.4} & \textbf{88.7} & \textbf{88.8} \\
\bottomrule
\end{tabular}
}
\end{table*}

\begin{table*}[ht]
\centering
\caption{Comparisons with CLIP-base fine-tuning methods on VLCS for single-source DG in terms of leave-all-but-one-domain-out performance with ResNet50 as the backbone. Bold denotes the best scores.}
\label{tab:single-DG_vlcs}
\resizebox{0.85\textwidth}{!}{
\begin{tabular}{
>{\centering\arraybackslash}m{3cm}|
>{\centering\arraybackslash}m{1.5cm}
>{\centering\arraybackslash}m{1.5cm}
>{\centering\arraybackslash}m{1.5cm}
>{\centering\arraybackslash}m{1.5cm}
>{\centering\arraybackslash}m{1.5cm}
>{\centering\arraybackslash}m{1.2cm}}
\toprule
Method & Caltech & LableMe & Pascal & Sun & Avg \\
\midrule    
Lin. Prob.~\cite{radford2021learning} & 58.9 & 62.6 & 76.7 & 63.8 & 65.5 \\
CoOp~\cite{zhou2022learning} & \textbf{74.6} & 76.9 & 78.7 & 71.0 & 75.3 \\
CoCoOp~\cite{zhou2022conditional} & 70.0 & 58.7 & 80.3 & 63.8 & 68.2 \\
DPL~\cite{zhang2023domain} & 64.8 & 77.0 & \textbf{80.7} & 78.3 & 75.2 \\
SPG (ours) & 70.2 & \textbf{79.3} & 76.3 & \textbf{80.2} & \textbf{76.5} \\
\bottomrule
\end{tabular}
}
\end{table*}

\begin{table*}[ht]
\centering
\caption{Comparisons with CLIP-base fine-tuning methods on OfficeHome for single-source DG in terms of leave-all-but-one-domain-out performance with ResNet50 as the backbone. Bold denotes the best scores.}
\label{tab:single-DG_offcehome}
\resizebox{0.85\textwidth}{!}{
\begin{tabular}{
>{\centering\arraybackslash}m{3cm}|
>{\centering\arraybackslash}m{1.5cm}
>{\centering\arraybackslash}m{1.5cm}
>{\centering\arraybackslash}m{1.5cm}
>{\centering\arraybackslash}m{1.5cm}
>{\centering\arraybackslash}m{1.5cm}
>{\centering\arraybackslash}m{1.2cm}}
\toprule
Method & Art & Clipart & Product & Real & Avg \\
\midrule    
Lin. Prob.~\cite{radford2021learning} & 36.8 & 42.1 & 50.8 & 55.8 & 46.4 \\
CoOp~\cite{zhou2022learning} & 71.3 & 75.6 & 65.8 & \textbf{70.2} & 70.7 \\
CoCoOp~\cite{zhou2022conditional} & 72.0 & \textbf{75.7} & 65.1 & 69.7 & 70.6 \\
DPL~\cite{zhang2023domain} & 72.0 & 75.3 & 65.7 & 70.1 & 70.8 \\
SPG (ours) & \textbf{72.3} & 74.8 & \textbf{66.1} & \textbf{70.2} & \textbf{70.9} \\
\bottomrule
\end{tabular}
}
\end{table*}

\begin{table*}[ht]
\centering
\caption{Comparisons with CLIP-base fine-tuning methods on TerraIncognita for single-source DG in terms of leave-all-but-one-domain-out performance with ResNet50 as the backbone. Bold denotes the best scores.}
\label{tab:single-DG_terra}
\resizebox{0.85\textwidth}{!}{
\begin{tabular}{
>{\centering\arraybackslash}m{3cm}|
>{\centering\arraybackslash}m{1.5cm}
>{\centering\arraybackslash}m{1.5cm}
>{\centering\arraybackslash}m{1.5cm}
>{\centering\arraybackslash}m{1.5cm}
>{\centering\arraybackslash}m{1.5cm}
>{\centering\arraybackslash}m{1.2cm}}
\toprule
Method & Location38 & Location43 & Location46 & Location100 & Avg \\
\midrule    
Lin. Prob.~\cite{radford2021learning} & \textbf{29.6} & 16.0 & 18.3 & 29.1 & 23.3 \\
CoOp~\cite{zhou2022learning} & 23.7 & \textbf{39.2} & 40.8 & 19.2 & 30.7 \\
CoCoOp~\cite{zhou2022conditional} & 22.8 & 27.5 & 34.0 & 18.1 & 25.6 \\
DPL~\cite{zhang2023domain} & 23.2 & 32.9 & 28.5 & 29.1 & 28.4 \\
SPG (ours) & 21.5 & 30.3 & \textbf{40.8} & \textbf{36.5} & \textbf{32.3} \\
\bottomrule
\end{tabular}
}
\end{table*}

\begin{table*}[ht]
\centering
\caption{Comparisons with CLIP-base fine-tuning methods on DomainNet for single-source DG in terms of leave-all-but-one-domain-out performance with ResNet50 as the backbone. Bold denotes the best scores.}
\label{tab:single-DG_domainnet}
\resizebox{\textwidth}{!}{
\begin{tabular}{
>{\centering\arraybackslash}m{3cm}|
>{\centering\arraybackslash}m{1.5cm}
>{\centering\arraybackslash}m{1.5cm}
>{\centering\arraybackslash}m{1.5cm}
>{\centering\arraybackslash}m{1.5cm}
>{\centering\arraybackslash}m{1.5cm}
>{\centering\arraybackslash}m{1.5cm}
>{\centering\arraybackslash}m{1.5cm}
>{\centering\arraybackslash}m{1.2cm}}
\toprule
Method & Clipart & Infograph & Painting & Quickdraw & Real & Sketch & Avg \\
\midrule    
Lin. Prob.~\cite{radford2021learning} & 4.2 & 3.3 & 6.6 & 2.7 & 16.4 & 4.4 & 6.3 \\
CoOp~\cite{zhou2022learning} & \textbf{47.2} & 46.6 & 45.3 & 48.0 & 35.0 & 47.1 & 44.9 \\
CoCoOp~\cite{zhou2022conditional} & 46.8 & \textbf{48.0} & \textbf{46.3} & 50.4 & 34.4 & 47.1 & 45.5 \\
DPL~\cite{zhang2023domain} & 46.6 & 47.5 & 45.2 & 40.6 & 35.6 & \textbf{47.3} & 43.8 \\
SPG (ours) & 44.0 & 45.4 & 45.2 & \textbf{55.6} & \textbf{36.6} & 46.8 & \textbf{45.6} \\
\bottomrule
\end{tabular}
}
\end{table*}

\clearpage

\subsection{Ablation Study}

\textbf{Component ablation.} The domain prompt label and generative model are indispensable components of our SPG method and cannot be directly removed, but can be replaced.
For domain prompt label, We replace the text prompt used in our work with VP~\cite{bahng2022exploring}.
For backbone, we replace the CGAN~\cite{mirza2014conditional} with an MLP. Additional domain label and backbone ablations are as follows.

\begin{table*}[htbp]
\centering
\caption{Component ablation for multi-source DG in terms of mean leave-one-domain-out performance with ResNet50 as the backbone. Bold denotes the best scores.}
\begin{tabular}{
>{\centering\arraybackslash}m{2.5cm}|
>{\centering\arraybackslash}m{1.5cm}
>{\centering\arraybackslash}m{1.5cm}
>{\centering\arraybackslash}m{1.5cm}
>{\centering\arraybackslash}m{1.5cm}
>{\centering\arraybackslash}m{1.5cm}
>{\centering\arraybackslash}m{1.5cm}
}
\toprule
Examples & PACS & VLCS & O.H. & TerraInc. & Do.Net & Avg \\
\midrule
w/ VP & 89.4 & 79.8 & 67.8 & 19.1 &	44.7 & 60.2      \\
w/ MLP & 90.4 & 79.6 & 72.3 & 31.7 & 47.8 & 64.4       \\
SPG (Ours) & \textbf{92.8} & \textbf{84.0} & \textbf{73.8} & \textbf{37.5} & \textbf{50.1} & \textbf{67.6}   \\
\bottomrule
\end{tabular}
\label{tab:abaltion_component}
\end{table*}

\end{document}